\documentclass[11pt]{article}
\usepackage[margin=0.9in]{geometry}

\usepackage[round]{natbib}
\usepackage{amsmath,amssymb,amsthm}
\usepackage{bm}
\usepackage{xcolor}
\usepackage{caption}

\usepackage{shortcuts} % à remettre seulement si le .sty existe

\usepackage[titlenumbered,linesnumbered,ruled,noend]{algorithm2e}

\SetCommentSty{mycommfont}
\SetEndCharOfAlgoLine{}

\usepackage[textsize=tiny]{todonotes}
\usepackage{mdframed}

\newcounter{boxcounter}[section]

\newenvironment{memobox}[1]{%
  \refstepcounter{boxcounter}%
  \begin{mdframed}[
    linewidth=0.5pt,
    linecolor=black,
    innertopmargin=8pt,
    innerbottommargin=8pt,
    innerleftmargin=10pt,
    innerrightmargin=10pt,
  ]%
  \noindent\textbf{A bit of pain: #1}\par\medskip
}{%
  \end{mdframed}%
}

\usepackage{amsmath,amssymb,mathtools}

\title{Notes on generative modeling: flow matching, diffusion, optimal transport and Schrödinger bridge}
\author{
Titouan Vayer\\
\small Inria, Rennes, France.
}

\date{}

\begin{document}

\maketitle

\begin{abstract}
These notes recapitulate the high level mathematical 
principles behind different techniques for generative modeling.
I show the connections between optimal transport and standard techniques 
such as Schrödinger bridge and flow matching.
\end{abstract}

\section{Introduction}

I made these notes for the Peyresq 2026 summer school.
The goal is not to be perfectly self-contained, nor perfectly rigorous, but to convey the essential ideas, in my view.
There are many good monographs on this topic, such as
\citet{pierret2026flow,peyré2026optimaldiffusiontransportsmachine},
and I highly recommend this
\href{https://neurips.cc/virtual/2024/invited-talk/101133}{talk}, this blog post \citet{gagneux2025a}, 
or the \href{https://vdeborto.github.io/project/generative_modeling/session_5.pdf}{slides} by Valentin de Bortoli on the Schrödinger bridge.
These notes reflect my personal perspective on the topic; this is far from being the only way to think about these methods.
I have added boxes throughout these notes: the reader may skip them on a first reading, as they are intended for a deeper understanding of the mathematical objects involved.

\paragraph{Notations.} The set of probability distribution on a space $X$ is denoted $\Pcal(X)$ and the set of positive finite measures $\Mcal_+(X)$. The set of coupling between two probability distributions $\pi_0, \pi_1$ is noted $C(\pi_0, \pi_1)$. For a map $T, \ T\# \pi_0$ denotes the push-forward of $\pi_0$ by $T$, when $X_0 \sim \pi_0, \ T(X_0) \sim T\#\pi_0$, that is $T\#\pi_0 = \operatorname{Law}(T(X_0))$ when $X_0 \sim \pi_0$.

\section{The continuity equation}

The ultimate goal of generative modeling is to generate samples from a distribution $\pi_1$.
One approach is to start from a simple, easy-to-sample distribution $\pi_0$, draw samples from it, and then use some strategy to obtain a sample $X_1$ from another distribution $\pi_1$.
In practice, this can be achieved by using data points $x_1, \cdots, x_n$ already drawn from $\pi_1$.

The starting point of these notes is the continuity equation, which describes how a probability distribution evolves over time.
Let $(p_t)_{t \in [0,1]}$ be sequence of probability distribution such that $p_t \in \Pcal(\R^d)$.
Let $u: \R^d \times [0,1] \to \R^d$ such that for each $t, u_t:\R^d \to \R^d$ is a vector field which indicates how the mass ``moves'' at a point $x$.
The continuity equation writes
\begin{equation}
    \label{eq:continuity_equation}
    \tag{CE}
    \begin{split}
        &\partial_t p_t + \diver(u_t \cdot p_t) = 0 \\
        % &p_0 = \pi_0 \text{ and } p_1 = \pi_1\,.
    \end{split}
\end{equation}
How to interpret \eqref{eq:continuity_equation}? $\partial_t p_t(x)$ is the variation of the density during time at $x$, and for $v: \R^d \to \R^d, v(x) = (v_1(x), \cdots, v_d(x))$ the quantity $\diver(v)(x) = \sum_{i=1}^d \frac{\partial v_i}{\partial x_i} (x)$ is the trace of the Jacobian of $v$. 
The divergence is nicely illustrated in \href{https://thehighergeometer.wordpress.com/2018/07/28/what-is-the-curl-of-a-vector-field-really/}{here} and gives the rate that the vector field alters the volume at around a point. 
Overall, \eqref{eq:continuity_equation} can be understood as the transport of some quantity of mass $p_t$ according to the vector field $u_t$.
\newpage
\begin{memobox}{weak form of the continuity equation}
We can generalize the continuity equation to probability distributions without a density, through what we call the weak formulation of the continuity equation.
To define it, we should view a probability distribution as something that \emph{measures} something (yes\ldots).

Precisely, a probability distribution $\mu$ can be characterized as an object that takes a real-valued function $\phi$ and outputs the real number $\int \phi(x) \dr \mu(x)$.
We say that $\mu$  \emph{acts} on test functions $\phi$ by \emph{duality}.
This is called a \emph{distribution}: it is a function that takes functions as input, is \emph{linear} in $\phi$ (and continuous), and outputs a scalar.
We will not be too formal about the regularity of these functions $\phi$; they are ``nice functions'' (compactly supported and infinitely differentiable).

So, in order to generalize to the case without a density, we should first view probability distributions as distributions, and then define all the relevant quantities.

    We begin by $\partial_t p_t$ that will denote the distribution that maps $\phi \to \frac{\dr }{\dr t}\int \phi(x) \dr p_t(x)$.
    When $p_t$ has density $f_t$, $\int \phi(x) \dr p_t(x) = \int \phi(x) f_t(x) \dr x$, so $\partial_t p_t$ is the object that maps $\phi \to \int \phi(x) (\frac{\dr }{\dr t} f_t(x)) \dr x$ which is what we want: $\partial_t p_t$ acts as $\partial_t f_t$ when it has a density.

    With the same idea we can define what is the derivative of a probability distribution, like $\partial_i \mu = \frac{\partial }{\partial x_i} \mu$ for $\mu \in \Pcal(\R^d)$.
    This will be the distribution that maps $\phi \to - \int [\partial_i \phi(x)] \dr \mu(x)$. Again definition is based on the fact that $\int \phi(x) (\partial_i f(x)) \dr x = - \int (\partial_i \phi(x)) f(x) \dr x$ using integration by parts and the fact that the test functions are ``nice''. So the derivative formulation matches the case when $\mu$ has a density.

Slightly more intricate, but again, by duality with test functions, and inspired by the density case, we can similarly define $\diver(u_t p_t)$ (the divergence of a probability distribution times a vector field) as the distribution $\phi \mapsto - \int \langle \nabla \phi(x), u_t(x)\rangle \dr p_t(x)$.
This is based on the following calculation: for a vector field $v: \R^d \to \R^d$ and a test function $\phi$, we have, by integration by parts,
\begin{equation}
\begin{split}
\int_\Omega \diver(v)(x)  \phi(x) \dr x &= \int_\Omega \sum_{i=1}^{d} \partial_i v_i(x) \phi(x) \dr x = -   \int \sum_{i} v_i(x) \partial_i \phi(x) \dr x \\
&=- \int \langle v(x), \nabla \phi(x) \rangle \dr x\,.
\end{split}
\end{equation}
So the divergence of a vector field $v$ can be seen as the distribution $\phi \to - \int \langle v(x), \nabla \phi(x) \rangle \dr x$. Now if $v = f u$ with $f: \R^d \to \R$ (such as a density), then $\int \langle v(x), \nabla \phi(x) \rangle \dr x = \int \langle u(x), \nabla \phi(x) \rangle f(x) \dr x$, justifying the definition of $\diver(u_t p_t)$ as the distribution $\phi \to - \int \langle \nabla \phi(x), u_t(x)\rangle \dr p_t(x)$. With similar calculus we can also check that this definition is compatible with the standard definition of the divergence of vector fields i.e. $\diver(uf)(x)= \sum_{i=1}^{d} \partial_i(u_i f)(x)$ (it defines the same distribution).
    
    Overall \eqref{eq:continuity_equation} ca be written, in what we call the \emph{weak form}, as an equality in the distributional sense:
    \begin{equation}
        \forall \text{ nice } \phi, \frac{\dr }{\dr t}\int \phi(x) \dr p_t(x) - \int \langle \nabla \phi(x), u_t(x)\rangle \dr p_t(x) = 0\,.
    \end{equation}
\end{memobox}

\begin{figure}[t]
    \centering
    \includegraphics[width=\linewidth]{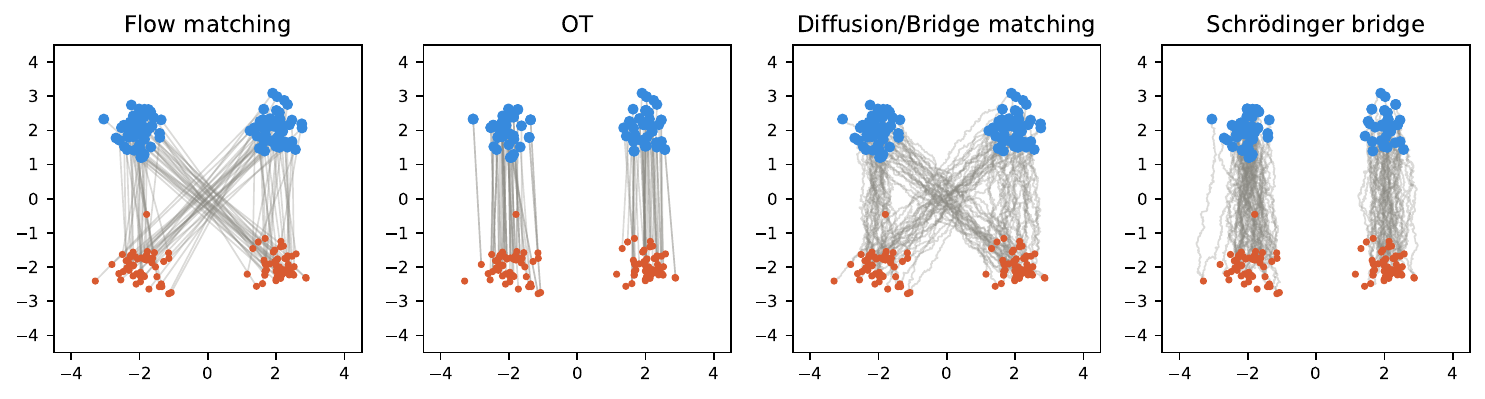}
    \caption{\label{fig:solvers} The different trajectories between $\pi_0, \pi_1$ that are obtained as $X_t = (1-t)X_0 + t X_1 + \sqrt{\varepsilon t (1-t)}Z$ with $(X_0, X_1)$ following a coupling $\gamma$ and $Z \sim \Ncal(0,I_d), \ \varepsilon \geq 0$. (Left) $\gamma$ independent coupling, $\varepsilon = 0$, (middle left) $\gamma$ is optimal transport coupling $\varepsilon = 0$, (middle right) $\varepsilon > 0, \gamma$ independent coupling, (right) $\varepsilon > 0$ and $\gamma$ entropic optimal transport coupling.}
\end{figure}

\paragraph{What if we have found $(u,p)$ that satisfies \eqref{eq:continuity_equation}?} In this case we can do generative modeling! Indeed, consider what we call a \emph{flow} $\Phi_t: \R^d \to \R^d$ that solves, for $t\in [0,1]$, the equation
\begin{equation}
    \label{eq:ode}
    \tag{ODE}
    \begin{split}
        &\frac{\dr \Phi_t(x)}{\dr t} = u_t(\Phi_t(x)) \\
        &\Phi_0(x) = x\,.
    \end{split}
\end{equation}
Then we can define $\tilde{p}_t = \Phi_t \# \pi_0$, it satisfies $\tilde{p}_0 = \pi_0$. 
Moreover, we can show that $(u, \tilde{p})$ also satisfies\footnote{Take $\phi$ then $\partial_t \int \phi \dr \tilde{p}_t = \partial_t \int \phi(\Phi_t(x)) \dr \pi_0(x) = \int \langle \nabla \phi(\Phi_t(x)), \partial_t \Phi_t(x) \rangle \dr \pi_0(x) =  \int \langle \nabla \phi(\Phi_t(x)), u_t(\Phi_t(x)) \rangle \dr \pi_0(x)= \int \langle \nabla \phi(x), u_t(x) \rangle \dr p_t(x)$, using chain rule, change of variable, and the hypothesis on $Z_t$.} \eqref{eq:continuity_equation}. 
Under some regularity conditions on $u$ we know that the $p$ such that $(u,p)$ satisfies \eqref{eq:continuity_equation} is unique, determined by the initial condition $p_0$ (see e.g. \citealt[Theorem 4.4.]{Santambrogio2015OptimalTF}). 
Thus, if $p_0 = \pi_0$ we have $\tilde{p}_t = p_t$ for any $t \in [0,1]$. If also $p_1 = \pi_1$ then
\begin{equation}
    \tilde{p}_1 = p_1 \implies \Phi_1 \# \pi_0 = \pi_1\,.
\end{equation}
Consequently, we have found a map $\Phi_1: \R^d \to \R^d$ that pushes the distribution $\pi_0$ forward to the distribution $\pi_1$!
So if $\pi_0$ is Gaussian, then to generate a sample according to $\pi_1$ we simply sample $x_0 \sim \pi_0$, and $\Phi_1(x_0)$ will be distributed as $\pi_1$.

Concretely, if we have found $(u, p)$ solving the continuity equation with $p_0 = \pi_0$ and $p_1 = \pi_1$, we solve an ODE and obtain a pushforward of $\pi_0$ to $\pi_1$.

In practice, to solve the ODE one can use any favorite solver, such as Euler's method: starting from $\Phi_0 = x$, discretize \eqref{eq:ode} with $t_k = k \Delta t$ (e.g.\ $\Delta t = 1/N$) and iterate
\begin{equation}
    \Phi_{k+1} = \Phi_k + u_{t_k}(\Phi_k) \, \Delta t\,,
\end{equation}
which yields an approximate solution $\Phi_{t}(x)$ of the ODE.

\paragraph{Which pair satisfies the continuity equation?}
The beauty of the flow matching model \citep{lipman2022flow,albergo2022building} is that it provides a pair $(u, p)$ satisfying \eqref{eq:continuity_equation} with $p_0 = \pi_0$ and $p_1 = \pi_1$, along with a way to approximate the velocity field $u$ with a neural network.

Consider $\gamma \in C(\pi_0, \pi_1)$ \textbf{any} coupling of $\pi_0$ and $\pi_1$. Then the pair
\begin{equation}
    \label{eq:solution_of_fm}
    \begin{split}
    &p_t = \operatorname{Law}((1-t) X_0 + t X_1), \text{ where } (X_0, X_1) \sim \gamma\,, \\
    &u_t(x) = \E_{(X_0, X_1)\sim \gamma}[X_1 - X_0 \mid (1-t)X_0+t X_1 = x]\,,
    \end{split}
\end{equation}
satisfies the continuity equation with $p_0 = \pi_0$ and $p_1 = \pi_1$.
The good thing is that we can approximate $u_t$ using a neural network, exploiting the fact that the conditional expectation is the solution of an optimization problem: $u = (u_t)$ as defined above is the solution of
\begin{equation}
    \label{eq:eq_to_optim}
    \underset{v: \R^d \times [0, 1] \to \R^d}{\operatorname{inf}} \ \int_{[0,1]}\int_{\R^d} \|(x_1 - x_0) - v((1-t)x_0 + tx_1, t)\|^2 \dr \gamma(x_0,x_1) \dr t\,.
\end{equation}
Thus, use your favorite neural network $v_\theta$, optimize \eqref{eq:eq_to_optim} with your favorite optimizer, and solve the ODE with $v_\theta \approx u$.

\paragraph{Are all couplings created equal?}
The previous reasoning can be carried out for any coupling $\gamma$, but this would lead to different trajectories, as illustrated in Figure~\ref{fig:solvers}.
Consider the first two subplots: the first illustrates the independent coupling $\gamma = \pi_0 \otimes \pi_1$, and the second is obtained with the coupling that solves the OT problem
\begin{equation}
    \label{eq:ot}
    \tag{OT}
 \min_{\gamma \in C(\pi_0, \pi_1)} \ \E_{(X_0, X_1) \sim \gamma}[\|X_0 - X_1\|^2]\,.   
\end{equation}
The latter yields straighter trajectories when solving the ODE with the corresponding velocity field $u_\theta(x) \approx u_t(x) = \mathbb{E}\!\left[X_1 - X_0 \mid (1-t)X_0 + tX_1 = x\right]$. This has a practical impact, as it leads to an Euler scheme that converges in significantly fewer steps, thereby reducing the number of evaluations of $u_\theta$.
This is intuitive: for the independent coupling, since any point from the blue distribution is matched to all points of the red distribution, at $t=0$ the velocity averages half of $X_1 - X_0$ from the blob ``in front of it'', but also one half from the other blob, so that some trajectories of the ODE curve toward the middle instead of going straight. On the other hand, OT coupling only applies to blobs facing each other, which would result in straight trajectories.

\paragraph{Benamou--Brenier formula.}
Using the optimal transport coupling is actually equivalent to solving the following dynamical problem, owing to the celebrated Benamou--Brenier formula \citep{benamou2000computational},
\begin{equation}
    W_2^2(\pi_0, \pi_1) = \min_{\gamma \in C(\pi_0, \pi_1)} \ \E_{(X_0, X_1) \sim \gamma}[\|X_0 - X_1\|^2] = \min_{\begin{smallmatrix} (u,p) \text{ satisfying } \eqref{eq:continuity_equation} \\ p_0 = \pi_0, p_1 = \pi_1 \end{smallmatrix}} \ \int_{\R^d} \int_{[0,1]} \|u_t(x)\|_2^2 \dr p_t(x)\dr t\,.
\end{equation}
The relation between the optimizers is simply that the $(u, p)$ minimizing the expression above are those defined in \eqref{eq:solution_of_fm} with $\gamma$ being the optimal coupling minimizing $\E_{(X_0, X_1) \sim \gamma}[\|X_0 - X_1\|^2]$.
The interpretation is quite nice: among all probability paths and vector fields, optimal transport finds the pair with the smallest energy, subject to the continuity equation constraint.

\section{Entropic optimal transport}

To introduce Schrödinger bridge we will need entropic optimal transport theory, I only give below a brief introduction and refer to \citet[Chap. 4]{peyre2019computational} for a better description.

\subsection{The Kullback-Leiber divergence}

For two finite positive measures $\mu, \nu \in \Mcal_+(X)$ on a space $X$ (not necessarily probability distributions, $0 <\mu(X) <+\infty$ but may be not $=1$) we define the Kullback-Leiber divergence as 
\begin{equation}
    \KL(\mu|\nu) = \begin{cases} \int_X \log(\frac{\dr \mu}{\dr \nu}(x)) \dr \mu(x) - \mu(X) + \nu(X) & \text{if } \mu \ll \nu\\
        +\infty &\text{ otherwise}\,.
    \end{cases}
\end{equation}
The notation $\mu \ll \nu$ simply means that $\nu(A) = 0 \implies \mu(A) = 0$. We say that  $\mu$ is ``dominated'' by $\nu$, when $\nu$ vanishes so does\footnote{Think about the discrete case: $\mu = (\mu_i)_{i\in \integ{n}}, \mu_i \geq 0, \nu = (\nu_j)_{j \in \integ{n}}, \nu_i \geq 0$ then $\mu \ll \nu \iff \nu_i = 0 \implies \mu_i = 0$.} $\mu$.
When this holds we know, from the Radon-Nikodym theorem that $\mu$ admits a density with respect to $\nu$, that is $\mu(A) = \int_A f(x) \dr \nu(x)$ for some unique non-negative density function $f \geq 0$. 
In this case $f=\frac{\dr \mu}{\dr \nu}$ so that the Kullback-Leiber divergence also writes$\footnote{Again, in the discrete case, $\KL(\mu|\nu) = \sum_i \log(\mu_i/\nu_i) \mu_i - \mu_i + \nu_i$.}$ $\KL(\mu|\nu) = \int (f\log f - f + 1)\dr \nu$. 

From the convexity of $- \log x, \ x \log x - x + 1 \geq 0$, hence
\begin{equation*}
    \KL(\mu | \nu) = \int_X (f(x)\log f(x) - f(x) + 1)\dr \nu(x) \geq 0\,.
\end{equation*}
In fact, when $\mu,\nu$ are prob. distributions, we have that $\KL(\mu | \nu) = 0$ if and only if $\mu = \nu$, so $\KL$ is called a \emph{divergence} between probability distributions. 
 
\subsection{Entropic OT and the Sinkhorn algorithm}

\begin{figure}[t]
    \centering
    \includegraphics[width=0.9\linewidth]{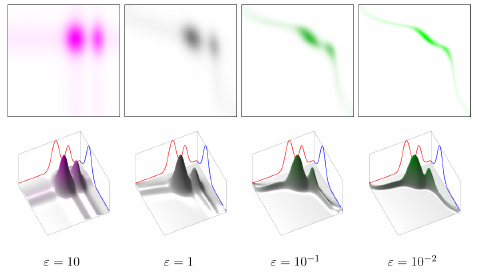}
    \caption{\label{fig:reg}Influence of regularization on the transport plan. Image taken from \cite{peyre2019computational}.}
\end{figure}
The entropic optimal transport problem (EOT) \citep{cuturi2013sinkhorn} between $\pi_0$ and $\pi_1$ writes, for\footnote{Of course, this can be defined for a cost other than the squared Euclidean cost.} $\varepsilon > 0$,
\begin{equation}
    \label{eq:entropic}
    \EOT_\varepsilon(\pi_0, \pi_1) = \min_{\gamma \in C(\pi_0, \pi_1)} \int \|x_0 - x_1\|^2 \dr \gamma(x_0, x_1) + \varepsilon \KL(\gamma | \pi_0 \otimes \pi_1)\,.
\end{equation}
\paragraph{First interpretation.} EOT looks for the coupling that minimizes the squared distances, while remaining close to the independent coupling $\pi_0 \otimes \pi_1$; this deviation is measured through the KL divergence term.
This term forces the coupling to remain spread out and regular, as the independent coupling is the one with the largest entropy.
As $\varepsilon \to \infty$, the penalization dominates and the optimal coupling becomes uniform and ``blurry''; conversely, as $\varepsilon \to 0$, the optimal coupling becomes sparser and sparser, and the minimum converges to the standard Wasserstein distance.
This influence is shown in Figure~\ref{fig:reg}.

\begin{remark}
    The KL term is always well-defined since any coupling $\gamma \in C(\pi_0, \pi_1)$ is absolutely continuous with respect to $\pi_0 \otimes \pi_1$, i.e., $\gamma \ll \pi_0 \otimes \pi_1$: in the discrete case, whenever there is no supply or demand between $i$ and $j$ (i.e., $\pi_0(i) = 0$ or $\pi_1(j) = 0$), the coupling satisfies $\gamma_{ij} = 0$, meaning no mass is transported between these points.
\end{remark}

\paragraph{Second interpretation.} Noting $C(x_0,x_1)$ the squared euclidean cost and $\langle C, \gamma\rangle = \int C \dr \gamma$, simple manipulations show that\footnote{The constant is equal to $ 1 - K_\varepsilon(\R^d \times \R^d)$}
\begin{equation}
    \langle C,  \gamma \rangle +\varepsilon \KL(\gamma| \pi_0 \otimes \pi_1) = \varepsilon \KL(\gamma | K_\varepsilon) + \operatorname{cte} \text{ where } \dr K_\varepsilon(x_0,x_1) = e^{- \frac{C(x_0,x_1)}{\varepsilon}} \dr (\pi_0\otimes \pi_1)(x_0, x_1)\,.
\end{equation}
Indeed,
\begin{equation}
    \begin{split}
        \KL(\gamma | K_\varepsilon)&= \int \log(\frac{\dr \gamma(x,y)}{\dr K_\varepsilon(x_0,x_1)}) \dr \gamma(x_0,x_1) - 1 + K_\varepsilon(\R^d \times \R^d)\\
        &= \int \log(\frac{\dr \gamma(x_0,x_1)}{\dr \pi_0 \otimes \pi_1(x_0,x_1)}\frac{\dr \pi_0 \otimes \pi_1(x_0,x_1)}{\dr K_\varepsilon(x_0,x_1)}) \dr \gamma(x_0,x_1) - 1 + K_\varepsilon(\R^d \times \R^d) \\
        &=\KL(\gamma | \pi_0 \otimes \pi_1) - \int \log(\frac{\dr K_\varepsilon(x_0,x_1)}{\dr \pi_0 \otimes \pi_1(x_0,x_1)})\dr \gamma(x_0,x_1) - 1 + K_\varepsilon(\R^d \times \R^d) \\
        &=\KL(\gamma | \pi_0 \otimes \pi_1) - \int \log(\exp(- C(x_0, x_1)/\varepsilon))\dr \gamma(x_0,x_1) - 1 + K_\varepsilon(\R^d \times \R^d) \\
        &=\KL(\gamma | \pi_0 \otimes \pi_1) + \int \frac{C(x_0, x_1)}{\varepsilon} \dr \gamma(x_0,x_1) + \operatorname{cte}\,.
    \end{split}
\end{equation}
In other words, since the constant term is not involved in the optimization, the EOT problem also looks for the coupling $\gamma$ that is closest to the measure $K_\varepsilon$, the latter being known as the Gibbs measure. In other words, the EOT problem is equivalent to
\begin{equation}
    \min_{\gamma \in C(\pi_0, \pi_1)} \ \KL(\gamma|K_\varepsilon)\,.
\end{equation}
One very important property is that \eqref{eq:entropic} is a strictly convex problem, so it admits a unique solution.
Using first-order conditions, we can show that $\gamma$ is optimal if and only if there exist functions $f, g > 0$ (the dual variables) such that
\begin{equation}
    \dr \gamma(x_0,x_1) = \exp(-C(x_0, x_1)/\varepsilon) f(x_0) g(x_1) \dr \pi_0(x_0) \dr\pi_1(x_1)\,.
\end{equation}
This shows that the optimal coupling is \emph{dense}: it is strictly positive everywhere, as opposed to the unregularized optimal transport.

More importantly, this solution can be computed using the Sinkhorn algorithm.
The idea is quite simple: to find the probability distribution closest to $K_\varepsilon$ with marginals $\pi_0$ and $\pi_1$, start from $K_\varepsilon$ and normalize the marginals alternately, so as to recover the correct marginals in the end.

Take the discrete case $K = (K_{ij})$ where $K_{ij} = \exp(-C_{ij}/\varepsilon)$ and $\gamma = (\gamma_{ij})$ is a matrix.
The Sinkhorn algorithm starts from $\gamma^{(0)} = K$ and alternatively computes 
\begin{equation}
    \begin{split}
    \gamma^{(2\ell)} &= \underset{\gamma \geq 0, \gamma 1 = \pi_0}{\operatorname{argmin}} \ \KL(\gamma | \gamma^{(2\ell-1)}) = \diag(\frac{\pi_0}{\gamma^{(2\ell-1)} 1}) \gamma^{(2\ell-1)} \\
    \gamma^{(2\ell+1)} &= \underset{\gamma \geq 0, \gamma^\top 1 = \pi_1}{\operatorname{argmin}} \ \KL(\gamma | \gamma^{(2\ell)}) =  \gamma^{(2\ell)} \diag(\frac{\pi_1}{{\gamma^{(2\ell)}}^\top 1})\,,
    \end{split}
\end{equation}
where division is meant pointwise. 
Practically, the iterates are defined equivalently using variable $u,v$ and involving the $K$ matrix, as bellow.
\begin{definition}
    The Sinkhorn algorithm starts from a strictly positive $v^0$ and iterates:
    \begin{align}
        u^{(\ell + 1)} &= \pi_0 / Kv^{(\ell)} \\
        v^{(\ell + 1)} &= \pi_1 / K^\top u^{(\ell + 1)}
    \end{align}
    where division is meant pointwise.
\end{definition}
These iterations are more natural as they correspond to the dual problem of EOT (that will be not described here).
An interpretation of this algorithm is also:
\begin{itemize}
    \item
    for a fixed $v$, the choice of $u$ that makes $\gamma = \diag(u) K \diag(v)$ satisfy $\diag(u) K \diag(v) 1 = \pi_0$ is $\pi_0/Kv$ (this is because $K \diag(v) 1 = Kv$).
    \item
    for a fixed $u$, the choice of $v$ that makes $\gamma = \diag(u) K \diag(v)$ satisfy $(\diag(u) M \diag(v))^\top 1 = \pi_1$ is $\pi_1 / K^\top u$.
\end{itemize}
So, Sinkhorn alternatively modifies $u$ and $v$ to satisfy the first and the second constraints.
Surprisingly, this works: Sinkhorn's algorithm converges to a solution of the matrix scaling problem, and hence can be used to solve EOT.

\section{Schrödinger bridge}

We will see two characterizations of the Schrödinger bridge \citep{schrodinger1931umkehrung,leonard2013survey}: both are equivalent and leads to the same solution under suitable choices. The first one is the dynamical definition.

\subsection{From continuity equation to Fokker-Planck equation}

We introduce a new equation, called the Fokker--Planck equation, which adds an extra term to the continuity equation and admits another nice interpretation.
The equation reads, for $\sigma_t > 0$,
\begin{equation}
    \label{eq:fokker_planck}
    \tag{FP}
    \begin{split}
        &\partial_t p_t + \diver(u_t \cdot p_t) = \frac{\sigma_t^2}{2} \Delta p_t \\
    \end{split}
\end{equation}
The term $\Delta p_t$ is the Laplacian of $p_t$: when $p_t$ has a density $f_t$, it can be understood as $\Delta p_t = \sum_{i=1}^{d} \frac{\partial^2 f_t}{\partial x_i^2}(x)$, and models the fact that the particles spread out.

\begin{memobox}{weak form of Fokker-Planck}
    As in the case of the continuity equation $\Delta p_t $ can be defined by duality with test functions $\phi$. $\Delta p_t $ is simply the distribution that maps $\phi \to \int \Delta \phi(x) \dr p_t(x)$. 
    So the weak form of the Fokker-Planck equation is 
    \begin{equation}
        \forall \text{ nice } \phi, \frac{\dr }{\dr t}\int \phi(x) \dr p_t(x) - \int \langle \nabla \phi(x), u_t(x)\rangle \dr p_t(x) = \frac{\sigma_t^2}{2} \int \Delta \phi(x) \dr p_t(x)\,.
    \end{equation}
\end{memobox}

Denoting by $\nabla \log p_t$ the \emph{score function}, we have a nice relation between the continuity equation and the Fokker--Planck equation:
\begin{equation}
    \label{eq:equivalence_fp_ce}
    (u_t,p_t) \text{ satisfies } \eqref{eq:fokker_planck} \iff \left(u_t - \frac{\sigma_t^2}{2}\nabla \log p_t, p_t\right) \text{ satisfies } \eqref{eq:continuity_equation}\,.
\end{equation}
The score can be understood as $\nabla \log f_t = \frac{\nabla f_t}{f_t}$ when $p_t$ has a density: it measures the variation of the density relative to its value.
The proof of \eqref{eq:equivalence_fp_ce} simply relies on the fact that\footnote{For the last equality, $\diver(\nabla p_t): \phi \mapsto - \sum_i \langle \partial_i p_t, \partial_i \phi\rangle = \sum_i \int \partial_{ii} \phi \, \dr p_t = \Delta p_t$.}
\begin{equation}
    \diver(\nabla \log p_t \cdot p_t) = \diver(\nabla p_t) = \Delta p_t\,.
\end{equation}

\begin{memobox}{score function for general probability distributions}
We can define the score function $\nabla \log \mu$ for a general probability distribution $\mu \in \Pcal(\R^d)$ by analogy with the density case. When it exists, the score is the vector field $\beta : \R^d \to \R^d$ such that $\nabla \mu = \beta \cdot \mu$.
However, this definition also requires us to define what the gradient of a probability distribution and $\beta \cdot \mu$ are.
For the first, $\nabla \mu = (\partial_1 \mu, \cdots, \partial_d \mu)$, where $\partial_i \mu$ is defined by duality as explained before: it is the distribution that maps $\phi \mapsto - \int (\partial_i \phi) \dr \mu$.
Writing $\beta(x) = (\beta_1(x), \cdots, \beta_d(x))$, the second term is defined via duality as $\beta \cdot \mu = (\beta_1 \cdot \mu, \cdots, \beta_d \cdot \mu)$, where $\beta_i \cdot \mu: \phi \mapsto \int \phi(x) \beta_i(x) \dr \mu(x)$.
\end{memobox}
This equivalence will be useful latter for interpreting the dynamical formulation of Schrödinger bridge. 

\subsection{From Fokker-Planck equation to SDE and diffusion/ bridge matching}

The relation between flow matching and the continuity equation has a direct counterpart for diffusion and bridge matching models (in the broad sense) \citep{song2020score,ho2020denoising,liu2022let,peluchetti2023non}.
Let $u_t$ be fixed, and consider a \emph{random} vector $X_t \in \R^d$.
Then $X_t$ satisfies the stochastic differential equation
\begin{equation}
    \tag{SDE}
    \label{eq:sde}
    \begin{split}
    &\dr X_t = u_t(X_t) \dr t + \sigma_t \dr B_t\\
    &X_0 \sim \pi_0\,.
    \end{split}
\end{equation}
if and only if the pair $(u, p)$ satisfies \eqref{eq:fokker_planck} with $p_t = \operatorname{Law}(X_t)$.
This result is based on Itô's formula, as described in the box below.

Now, how can we choose $u$ such that solving \eqref{eq:sde} yields $X_1 \sim \pi_1$?
From the earlier analysis we know that, for $\gamma \in C(\pi_0, \pi_1)$,
\begin{equation}
    \begin{split}
    &\tilde{p}_t = \operatorname{Law}((1-t) X_0 + t X_1), \text{ where } (X_0, X_1) \sim \gamma\,, \\
    &v_t(x) = \E_{(X_0, X_1)\sim \gamma}[X_1 - X_0 \mid (1-t)X_0+t X_1 = x]\,,
    \end{split}
\end{equation}
is such that $(v, \tilde{p})$ satisfies the continuity equation with $p_0 = \pi_0$ and $p_1 = \pi_1$.
Then, from \eqref{eq:equivalence_fp_ce}, we know that $\left(v_t + \frac{\sigma_t^2}{2}\nabla \log \tilde{p}_t, \tilde{p}_t\right)$ satisfies \eqref{eq:fokker_planck}, and by definition $\tilde{p}_0 = \pi_0$ and $\tilde{p}_1 = \pi_1$.
Now solve \eqref{eq:sde} with $u_t = v_t + \frac{\sigma_t^2}{2} \nabla \log \tilde{p}_t$ and define $p_t = \operatorname{Law}(X_t)$. Then $(u_t, p_t)$ also satisfies \eqref{eq:fokker_planck}. Since $\tilde{p}_0 = p_0$, by uniqueness (again, hiding the regularity details on the velocity field), $\tilde{p}_t = p_t$ for all $t$, and in particular $p_1 = \tilde{p}_1 = \pi_1$, so $X_1 \sim \pi_1$!

So in practice, if we can approximate $v_t + \frac{\sigma_t^2}{2} \nabla \log \tilde{p}_t$ with a neural network, we run \eqref{eq:sde} with this network (using a solver similar to Euler's method, but adding noise at each step), and obtain $X_1 \sim \pi_1$.
The good news is that when $\pi_0$ is Gaussian, Tweedie's formula shows that $\nabla \log \tilde{p}$ can also be written as a conditional expectation: we can apply the same trick to approximate it as a regression problem with a neural network.

\begin{memobox}{what is an SDE? the Itô formula? relations with Fokker-Planck?}
\eqref{eq:sde} is the stochastic equivalent of \eqref{eq:ode} and to understand this we need to define objects. First \eqref{eq:sde} is more a notation than an equation. It should be understood precisely as the random vector $X_t \in \R^d$ that satisfies, for $t \in [0,1]$,  
\begin{equation}
    X_t = X_0 + \int_{0}^{t} u_s(X_s) \dr s + \int_{0}^{t} \sigma_s \dr B_s\,.
\end{equation}
Then we need to understand what is $\int_{0}^{t} \sigma_s \dr B_s$. $B_t$ is called a standard Brownian motion, it satisfies  
\begin{equation}
    \begin{split}
     &B_0 = 0 \\
     &B_{t + \Delta t} - B_{t} \sim \Ncal(0, \Delta t)\,.
    \end{split}
\end{equation}
These properties are then used to define $\int_{0}^{t} \sigma_s \dr B_s$. First, this quantity is simply a random vector that will be defined implicitly as the limit of some random vectors. 
Precisely, for $0 < t_0 < \cdots < t_{n-1} < t$ a subdivision of $[0, t]$, we note $Z_n = \sum_{k=0}^{n-1} \sigma_{t_k} (B_{t_{k+1}} - B_{t_k})$. It is a random vector, and it converges to some $Z$ as $n \to +\infty$ in the sense $\E[\|Z_n - Z\|_2] \to 0$ and $\int_{0}^{t} \sigma_s \dr B_s$ is defined as this limit $Z \in \R^d$.

Now the Itô formula can be seen as a generalization of the chain rule for stochastic processes. Given $X_t$ that satisfies \eqref{eq:sde}, and a real value function $\phi: \R^d \to \R$ then $\phi(X_t)$ also satisfies a SDE given by 
\begin{equation}
    \label{eq:ito}
    \phi(X_t) = \phi(X_0) + \int_{0}^{t} \left(\langle u_s(X_s), \nabla \phi(X_s) \rangle +\frac{\sigma_s^2}{2} \Delta \phi(X_s) \right)\dr s + \int_{0}^{t} \langle  \nabla \phi(X_s), \sigma_s \dr B_s\rangle\,.
\end{equation}
The last term $\int_{0}^{t} \langle  \nabla \phi(X_s), \sigma_s \dr B_s\rangle$ is also a limit: similarly as before it is defined as the limit in $L_2$ of the random variable $\sum_{k=0}^{n-1} \sigma_{t_k} \langle \nabla \phi(X_{t_k}),  (B_{t_{k+1}} - B_{t_k})\rangle$.

Finally, I show that $(u,p)$ where $p_t = \operatorname{Law}(X_t)$ satisfies \eqref{eq:fokker_planck}.
Indeed, for any nice $\phi$ from \eqref{eq:ito} we get
\begin{equation}
    \E[\phi(X_t)] = \E[\phi(X_0)] + \E[\int_{0}^{t} \left(\langle u_s(X_s), \nabla \phi(X_s) \rangle +\frac{\sigma_s^2}{2} \Delta \phi(X_s) \right)\dr s] + \E[\int_{0}^{t} \langle  \nabla \phi(X_s), \sigma_s \dr B_s\rangle]\,.
\end{equation} 
However, the last term $\E[\int_{0}^{t} \langle  \nabla \phi(X_s), \sigma_s \dr B_s\rangle] = 0$ (it is defined as the limit in $L_2$ of some $Y_n$ as $n \to \infty$ where $\E[Y_n]=0$).
Deriving in $t$ the previous equation then gives
\begin{equation}
    \frac{\dr }{\dr t} \E[\phi(X_t)] = \E[\langle u_t(X_t), \nabla \phi(X_t) \rangle +\frac{\sigma_s^2}{2} \Delta \phi(X_t)]\,.
\end{equation}
Now by definition of $p_t$ 
\begin{equation}
    \E[\langle u_t(X_t), \nabla \phi(X_t) \rangle +\frac{\sigma_s^2}{2} \Delta \phi(X_t)] = \int \langle u_t(x), \nabla \phi(x) \rangle +\frac{\sigma_t^2}{2} \Delta \phi(x) \dr p_t(x)\,,
\end{equation}
and $\frac{\dr }{\dr t} \E[\phi(X_t)] = \frac{\dr }{\dr t} \int \phi(x) \dr p_t(x)$. Combining both, and we obtain exactly the weak form of the Fokker-Planck equation.

\end{memobox}

\subsection{First formulation of Schrödinger bridge}

The previous reasoning holds for any coupling, and again we may obtain suboptimal trajectories for diffusion, as illustrated in Figure~\ref{fig:solvers}, for the same reasons as in flow matching. The Schrödinger bridge will provide a tool for finding better trajectories in theory.

The first formulation of the Schrödinger bridge aims at choosing, among all possible solutions to the Fokker--Planck equation, the one that minimizes the kinetic energy appearing in the Benamou--Brenier formula.
In other words, the Schrödinger bridge aims at solving, with $\sigma_t = \sqrt{\varepsilon}$,
\begin{equation}
    \tag{SB-Dyn}
    \label{eq:schordinger_dyn}
\min_{\begin{smallmatrix} (u,p) \text{ satisfying } \eqref{eq:fokker_planck} \\ p_0 = \pi_0, \ p_1 = \pi_1 \end{smallmatrix}} \ \int_{\R^d} \int_{[0,1]} \|u_t(x)\|_2^2 \dr p_t(x)\dr t\,.
\end{equation}

We also have another formulation of \eqref{eq:schordinger_dyn} via the change of variables $v = u - \frac{\varepsilon}{2} \nabla \log p$.
Suppose $p_t \ll \lambda$ for some $\lambda$; it can then be shown that $\int_{\R^d} \int_{[0,1]} \langle \nabla \log p_t(x), v_t(x) \rangle \dr p_t(x)\dr t = \KL(\pi_0| \lambda) - \KL(\pi_1|\lambda)$, so this term does not depend on $v$. In this case, \eqref{eq:schordinger_dyn} is equivalent to the following minimization problem over the continuity equation:
\begin{equation}
    \label{eq:equiv_sb1}
    \min_{\begin{smallmatrix} (u,p) \text{ satisfying } \eqref{eq:continuity_equation} \\ p_0 = \pi_0, \ p_1 = \pi_1 \end{smallmatrix}} \ \int_{\R^d} \int_{[0,1]} \|u_t(x)\|_2^2 + \frac{\varepsilon^2}{4}\|\nabla \log p_t(x)\|_2^2\dr p_t(x)\dr t\,.
\end{equation}

Equation \eqref{eq:equiv_sb1} admits a nice interpretation. Defining the Fisher information
\begin{equation}
    I(p_t) = \int \|\nabla \log p_t(x)\|_2^2 \dr p_t(x)\,,
\end{equation}
we see that \eqref{eq:equiv_sb1} is the Benamou--Brenier formula penalized by $\frac{\varepsilon^2}{4} \int I(p_t) \dr t$.
This term \emph{discourages concentration}. Take for instance the Gaussian $p_t = \Ncal(0, \sigma_t^2)$; then $I(p_t) = \frac{1}{\sigma^2_t}$, so a very small variance induces a very large penalty.
The variance of $p_t$ is thus forced to remain large, keeping $p_t$ spread out and regular.
This is reminiscent of the $\KL(\gamma| \pi_0 \otimes \pi_1)$ term, and in fact the two are deeply related.

\subsection{A Kullback-Leiber formulation of Schrödinger bridge}

The connection with entropic optimal transport will become evident (I hope) when one looks at a certain $\KL$ minimization problem.
To this end, we introduce some notation: we say that $\Pbb$ is a \emph{path measure} when $\Pbb \in \Pcal(C([0,1], \R^d))$ is a probability distribution on the space of continuous functions from $[0,1]$ to $\R^d$.
Equivalently, $\Pbb$ is a path measure when $\Pbb = \operatorname{Law}(X)$ for some random continuous path $X \in \mathcal{C}([0,1], \mathbb{R}^d)$, i.e.\ $X_t \in \R^d$. In other words, $\Pbb$ is the law of the whole trajectory, encoding the joint distribution of all time snapshots simultaneously.

In the following, we denote by $(\Pbb)_t$ the marginal law at time $t$. More formally, it can be written as $(\Pbb)_t = e_t \# \Pbb$, where
\begin{equation}
    \begin{split}
    e_t: C([0,1], \R^d) &\to \R^d    \\
    X &\mapsto X_t\,.     
    \end{split}
\end{equation}
\begin{remark}
    $\Pbb$ differs from the collection $(p_t)_{t\in[0,1]}$, where $p_t = \operatorname{Law}(X_t) = (\Pbb)_t$: the latter are only the marginals, i.e.\ the snapshots at time $t$, whereas $\Pbb$ is the joint distribution encoding the interactions between all timesteps.
\end{remark}

Now let $\Qbb$ be a \emph{reference} path measure. The Schrödinger bridge problem is
\begin{equation}
    \label{eq:schordinger_kl}
    \tag{SB-KL}
    \min_{\begin{smallmatrix} \Pbb \in \Pcal(C([0,1], \R^d)) \\ (\Pbb)_0 = \pi_0, \ (\Pbb)_1 = \pi_1 \end{smallmatrix}} \ \KL(\Pbb |\Qbb)\,.
\end{equation}
Of course, this problem makes sense only when $\Pbb \ll \Qbb$. The interpretation is that we seek a path measure, describing the law of trajectories, that is closest to a given reference path measure, with endpoints fixed at $\pi_0$ and $\pi_1$.

\paragraph{The fundamental result.} Using the disintegration theorem, we have that \eqref{eq:schordinger_kl} is equivalent to a problem very similar to entropic optimal transport between the endpoints $\pi_0$ and $\pi_1$.
Indeed, denoting by $\Qbb_{0,1} = (e_0 \times e_1) \# \Qbb \in \Pcal(\R^d \times \R^d)$ the joint law of the endpoints of $\Qbb$, equation \eqref{eq:schordinger_kl} is equivalent to
\begin{equation}
    \label{eq:schordinger_kl_3}
    \min_{\gamma \in C(\pi_0, \pi_1)} \ \KL(\gamma |\Qbb_{0,1})\,.
\end{equation}
This is beautiful in a sense: one only needs to match the endpoints to recover the optimal trajectory.
The relation between the optimizers is the following: when $\gamma$ is optimal for \eqref{eq:schordinger_kl_3}, then the optimal path measure $\Pbb$ is given by
\begin{equation}
    \dr \Pbb\left((X_t)_{t\in [0,1]}\right) = \dr\Qbb\left( (X_t)_{t\in [0,1]} \mid X_0 = x_0, X_1 = x_1\right) \dr\gamma(x_0, x_1)\,,
\end{equation}
meaning that the optimal trajectories are obtained by first sampling $(x_0, x_1) \sim \gamma$, and then sampling a path from $\Qbb$ conditioned on its endpoints.
However, we are not yet at entropic optimal transport, which will be the subject of the next section.

\subsection{Schrödinger is entropic optimal transport?}

By carefully choosing the reference $\Qbb$, we can recover the EOT problem.
We consider $\Qbb$ the path measure associated to the SDE
\begin{equation}
    \begin{split}
    &\dr X_t = \sqrt{\varepsilon} \ \dr B_t \\
    &X_0 \sim \pi_0\,, 
    \end{split}
\end{equation}
i.e.\ $\Qbb$ is the joint law of the solution to this SDE. In this case $\Qbb_{0,1}$ is easy to compute, since $X_1 = X_0 + \int_{0}^{1} \sqrt{\varepsilon} \dr B_t = X_0 + \sqrt{\varepsilon}(B_1 - B_0)$. Thus $X_1 | X_0 = x_0 \sim \Ncal(x_0, \varepsilon I_d)$, so
\begin{equation*}
    \dr \Qbb_{0,1}(x_0, x_1) = \frac{1}{Z_\varepsilon} \exp\!\left(- \frac{\|x_0 - x_1\|^2}{2\varepsilon}\right)\dr \pi_0 (x_0) \dr x_1\,, \text{where } Z_\varepsilon = \int_{\R^d \times \R^d} \exp\!\left(- \frac{\|x_0 - x_1\|^2}{2\varepsilon}\right)\dr \pi_0 (x_0) \dr x_1\,.
\end{equation*}
This is close to the $K_{2\varepsilon}$ seen before,
\begin{equation}
    \dr K_\varepsilon(x_0,x_1) = e^{- \frac{C(x_0,x_1)}{\varepsilon}} \dr (\pi_0\otimes \pi_1)(x_0, x_1)\,,
\end{equation}
except for a different marginal $\dr x_1$ instead of $\dr \pi_1(x_1)$ and a normalization constant. We will see that we can recover exactly the right measure up to a constant that does not depend on the coupling.

We compute $\KL(\gamma | \Qbb_{0,1})$ for $\gamma$ with $\gamma_0 = \pi_0$ and $\gamma_1 = \pi_1$:
\begin{equation}
    \KL(\gamma | \Qbb_{0,1}) = \int \log \frac{\dr \gamma}{\dr \Qbb_{0,1}}(x_0, x_1) \dr \gamma(x_0, x_1)\,.
\end{equation}
Writing, for $\lambda$ the Lebesgue measure,
\begin{equation}
    \log \frac{\dr \gamma}{\dr \Qbb_{0,1}}(x_0,x_1) = \log \frac{\dr \gamma}{\dr (\pi_0 \otimes \lambda)}(x_0,x_1) + \log \frac{\dr (\pi_0 \otimes \lambda)}{\dr \Qbb_{0,1}}(x_0,x_1)\,.
\end{equation}
Since $\dr \Qbb_{0,1}(x_0,x_1) = \frac{1}{Z_\varepsilon} e^{-\frac{\|x_0-x_1\|^2}{2\varepsilon}} \dr\pi_0(x_0) \dr x_1$, we have
\begin{equation}
    \log \frac{\dr(\pi_0 \otimes \lambda)}{\dr \Qbb_{0,1}}(x_0,x_1) = \frac{\|x_0-x_1\|^2}{2\varepsilon} + \log Z_\varepsilon\,.
\end{equation}
Introducing $\pi_1$ in the first term:
\begin{equation}
    \log \frac{\dr \gamma}{\dr(\pi_0 \otimes \lambda)}(x_0,x_1) = \log \frac{\dr \gamma}{\dr(\pi_0 \otimes \pi_1)}(x_0,x_1) + \log \frac{\dr \pi_1}{\dr \lambda}(x_1)\,.
\end{equation}
Integrating against $\gamma$ and using $\gamma_1 = \pi_1$,
\begin{equation}
    \int \log \frac{\dr \pi_1}{\dr \lambda}(x_1) \dr \gamma(x_0,x_1) = \int \log \frac{\dr \pi_1}{\dr \lambda}(x_1) \dr \pi_1(x_1) = C_{\pi_1}\,,
\end{equation}
which is a constant with respect to $\gamma$. Therefore,
\begin{equation}
    \KL(\gamma | \Qbb_{0,1}) = \KL(\gamma | \pi_0 \otimes \pi_1) + \frac{1}{2\varepsilon} \int \|x_0 - x_1\|^2 \dr \gamma(x_0,x_1) + C_{\pi_1} + \log Z_\varepsilon\,.
\end{equation}
Since $C_{\pi_1} + \log Z_\varepsilon$ does not depend on $\gamma$, minimizing $\KL(\gamma \| \Qbb_{0,1})$ over couplings $\gamma$ with $\gamma_0 = \pi_0$ and $\gamma_1 = \pi_1$ is equivalent to minimizing
\begin{equation}
    2\varepsilon\KL(\gamma | \pi_0 \otimes \pi_1) + \int \|x_0 - x_1\|^2 \dr \gamma(x_0,x_1)\,,
\end{equation}
which is exactly the EOT problem with cost $C(x_0,x_1) = \|x_0-x_1\|^2$ and parameter $2\varepsilon$.

Now, what about the dynamical formulation \eqref{eq:schordinger_dyn}? Here something called \emph{Girsanov theorem} \citep{follmer2006random,oksendal2003stochastic} implies that for any $\Pbb \ll \Qbb$ with $\Qbb$ as defined above, there exists a \emph{random} vector field $u_t \in \R^d$ such that $\Pbb$ is the path measure associated to the SDE $\dr X_t = \sqrt{\varepsilon}\, u_t \dr t + \sqrt{\varepsilon}\, \dr B_t$. Moreover,
\begin{equation}
    \KL(\Pbb|\Qbb) = \frac{1}{2}\int_{0}^{1} \E[\|u_t(X_t)\|_2^2] \dr t = \frac{1}{2}\int_{0}^{1} \int_{\R^d}\|u_t(x)\|_2^2\dr p_t(x) \dr t\,,
\end{equation}
with $p_t = \operatorname{Law}(X_t)$. Thus, minimizing the $\KL$ over all path measures with endpoints $\pi_0$ and $\pi_1$ is exactly solving \eqref{eq:schordinger_dyn}!
Overall, we have the following result.
\begin{theorem}
    With $\Qbb$ defined above and assuming $\pi_1 \ll \lambda$ the Lebesgue measure, the Schrödinger bridge problems \eqref{eq:schordinger_dyn} and \eqref{eq:schordinger_kl} and the entropic optimal transport problem $\EOT_{2\varepsilon}(\pi_0, \pi_1)$ are all equivalent.
\end{theorem}

Suppose that we can compute $\gamma$, the optimal coupling associated to $\EOT_{2\varepsilon}(\pi_0, \pi_1)$ (which we cannot in practice, since we do not have access to $\pi_1$ directly).
For each $(x_0, x_1) \sim \gamma$, an optimal trajectory according to the Schrödinger bridge follows the SDE
\begin{equation}
    \begin{split}
    &\dr X_t = \frac{x_1 - X_t}{1-t} \dr t + \sqrt{\varepsilon} \dr B_t \\
    &X_0 = x_0\,,         
    \end{split}
\end{equation}
which corresponds to $\Qbb(\cdot|X_0= x_0, X_1=x_1)$.
This process is the Brownian bridge starting at $x_0$ and ending at $x_1$ at time $t=1$, since
\begin{equation}
    X_t = (1-t)x_0 + t x_1 + \sqrt{\varepsilon}(1-t)\int_{0}^t \frac{\dr B_s}{1-s} = (1-t)x_0 + t x_1 + \sqrt{\varepsilon\,t(1-t)}\, Z\,, \quad \text{where } Z \sim \Ncal(0, I_d)\,.
\end{equation}
We used this to compute the trajectories in Figure~\ref{fig:solvers}, where $\pi_0$ and $\pi_1$ are known explicitly as discrete distributions in this small example.
In practice, this SDE requires to sample $x_1$, and we do not know how to do it (this is our goal). However, a fundamental results states that the SDE 
\begin{equation}
    \begin{split}
    &\dr X_t = \frac{\E_{\gamma}[X_1|X_t] - X_t}{1-t} \dr t + \sqrt{\varepsilon} \dr B_t\,,        
    \end{split}
\end{equation} 
has the same marginal distribution (this is called Markovian projection). Consequently, as before, a neural network can be used to approximate the $u_t$ defined above.

\section{Conclusion}

We have seen in these notes that the Schrödinger bridge and entropic optimal transport are closely related, but I have not really discussed algorithms to compute them in the general case.
To solve the Schrödinger bridge, one can use Sinkhorn between $\pi_0$ and $\pi_1$, but in practice we do not know $\pi_1$, we only have samples from it.
One algorithm consists in alternating forward process followed by a backward one, and  obtaining a new coupling at each step, used to train neural network. These iterations are exactly a noisy version of rectified flow \citep{shi2023diffusion, liu2022flow}.
What is particularly elegant is that iterating this process is guaranteed to converge to the true entropic optimal coupling, something that does not hold for standard rectified flow, which does not converge to the optimal transport coupling.

\bibliographystyle{unsrtnat}
\bibliography{references}

\end{document}